\documentclass[10pt,twocolumn,letterpaper]{article}

\usepackage{iccv}
\usepackage{times}
\usepackage{epsfig}
\usepackage{graphicx}
\usepackage{amsmath}
\usepackage{amssymb}
\usepackage{booktabs}
\usepackage{graphicx}
\usepackage{caption}
\usepackage{algorithm,algpseudocode}
\usepackage{algorithmicx}
\usepackage{makecell}
\usepackage[accsupp]{axessibility}

\usepackage{authblk}

\usepackage[pagebackref=true,breaklinks=true,letterpaper=true,colorlinks,bookmarks=false]{hyperref}

\iccvfinalcopy 

\def\iccvPaperID{12241} 

\ificcvfinal\fi

\begin{document}

\title{General Image-to-Image Translation with\\ One-Shot Image Guidance}


\author{Bin Cheng$^{*}$}
\author{Zuhao Liu$^{*}$}
\author{Yunbo Peng}
\author{Yue Lin}

\affil{NetEase Games AI Lab \\{\tt\small \{chengbin04, liuzuhao, gzpengyunbo, gzlinyue\}@corp.netease.com}}

\twocolumn[{%
\renewcommand\twocolumn[1][]{#1}%
\maketitle
\begin{center}
    \centering
    
    \captionsetup{type=figure}
    \includegraphics[width=\linewidth]{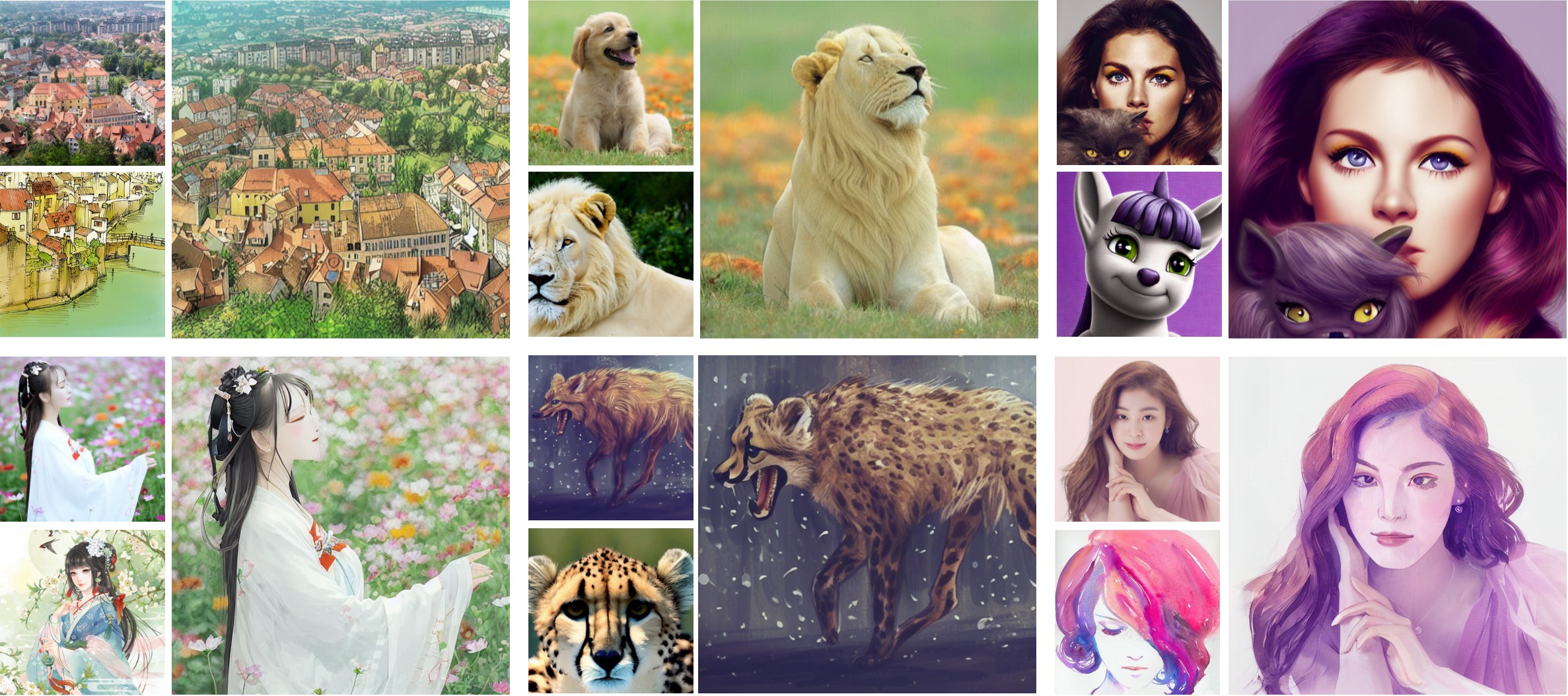}
    \captionof{figure}{The image-to-image translation exhibition of the proposed visual concept translator (VCT). For each image group, the upper-left part is the source image, the lower-left part is the reference image, and the right part is the model output (target image).}
    
\end{center}
}]

\ificcvfinal\thispagestyle{empty}\fi

\renewcommand{\thefootnote}{\fnsymbol{footnote}}
\footnotetext[1]{The first two authors contributed equally to this work.}

\begin{abstract}
Large-scale text-to-image models pre-trained on massive text-image pairs show excellent performance in image synthesis recently. However, image can provide more intuitive visual concepts than plain text. People may ask: how can we integrate the desired visual concept into an existing image, such as our portrait? Current methods are inadequate in meeting this demand as they lack the ability to preserve content or translate visual concepts effectively. Inspired by this, we propose a novel framework named visual concept translator (VCT) with the ability to preserve content in the source image and translate the visual concepts guided by a single reference image. The proposed VCT contains a content-concept inversion (CCI) process to extract contents and concepts, and a content-concept fusion (CCF) process to gather the extracted information to obtain the target image. Given only one reference image, the proposed VCT can complete a wide range of general image-to-image translation tasks with excellent results. Extensive experiments are conducted to prove the superiority and effectiveness of the proposed methods. Codes are available at https://github.com/CrystalNeuro/visual-concept-translator.

\end{abstract}

\section{Introduction}

Image-to-image translation (I2I) task aims to learn a conditional generation function that translates images from source to target domain with source content preserved and target concept transferred\cite{richardson2021encoding, wang2018high}. General I2I can complete a wide range of applications without dedicated model design or training from scratch \cite{wang2022pretraining}. 
Traditionally, generative adversarial networks (GAN) or normalizing flow \cite{grover2020alignflow} are mainly applied to I2I tasks \cite{karras2019style, karras2019style, richardson2021encoding, collins2020editing}. However, these methods suffer from the problem of lacking adaptability \cite{su2022dual}. The model trained in one source-target dataset cannot adapt to another one, so they fail to work in the scenario of general I2I. 

Diffusion-based image synthesis has been developed rapidly in recent years due to the application of large-scale models \cite{rombach2022high, saharia2022photorealistic, ramesh2022hierarchical}. Their strength is using a large number of image-text pairs for model training, so diverse images can be generated by sampling in the latent space guided by a specific text prompt. 
However, in our daily life, we accept massive visual signals containing abundant visual concepts. These visual concepts are difficult to describe in plain text just as the adage ``A picture is worth a thousand words".
In addition, I2I guided by reference images has wide applications including game production, artistic creation, and virtual reality. Therefore, research on image-guided I2I contains great potential in the computer vision community. 

Several works try to extract visual information from images with the desired concepts. Specifically, \cite{gal2022image} proposes a technique named textual inversion (TI) which freezes the model and learns a text embedding to represent the visual concepts. On the basis of TI, DreamBooth \cite{ruiz2022dreambooth} and Imagic \cite{kawar2022imagic} are proposed to alleviate overfitting caused by model fine-tuning.
The above methods are under the few-shot setting but sometimes collecting several related images containing the same concept is difficult. To address this problem, \cite{dong2022dreamartist} proposes to use both positive and negative text embedding to fit the one-shot setting. However, these methods cannot be directly used in I2I tasks because they cannot preserve the content in the source image. 

In order to preserve the source contents, the recently proposed DDIM inversion \cite{dhariwal2021diffusion, song2020denoising} finds out the deterministic noise along the reverse direction of the diffusion backward process. Then, some studies\cite{mokady2022null, hertz2022prompt} further apply and improve the DDIM inversion to text-guided image editing. However, these methods are text-conditional so they fail to understand the visual concepts from reference images.
Alternately, some works \cite{yang2022paint, su2022dual} try to connect the source and target domain with image condition, but their models are task-specific so they cannot be used in general I2I.

In this paper, to complete the general I2I tasks guided by reference images, we propose a novel framework named visual concept translator (VCT) with the ability to preserve content in the source image and translate the visual concepts with a single reference image. The proposed VCT solves the image-guided I2I by two processes named content-concept inversion (CCI) and content-concept fusion (CCF). The CCI process extracts contents and concepts from source and reference images through pivot turning inversion and multi-concept inversion, The CCF process employs a dual-stream denoising architecture to gather the extracted information to obtain the target image. Given only one reference image, the proposed VCT can complete a wide range of general image-to-image translation tasks with excellent results.
Extensive experiments including massive tasks of general I2I and style transfer are conducted for model evaluation.

In summary, our contributions are as follows

(1) We propose a novel framework named visual concept translator (VCT). Given only a single reference image, VCT can complete the general I2I tasks with the ability to preserve content in the source image and translate the visual concepts.

(2) We propose a content-concept inversion (CCI) to extract contents and concepts with pivot turning inversion and multi-concept inversion. We also propose a content-concept fusion (CCF) process to gather the extracted information with a dual-stream denoising architecture.

(3) Extensive experiments including massive tasks of general I2I and style transfer are conducted for model evaluation. The generation results show the high superiority and effectiveness of the proposed methods.

\begin{figure*}[t]
\begin{center}
\includegraphics[width=\linewidth]{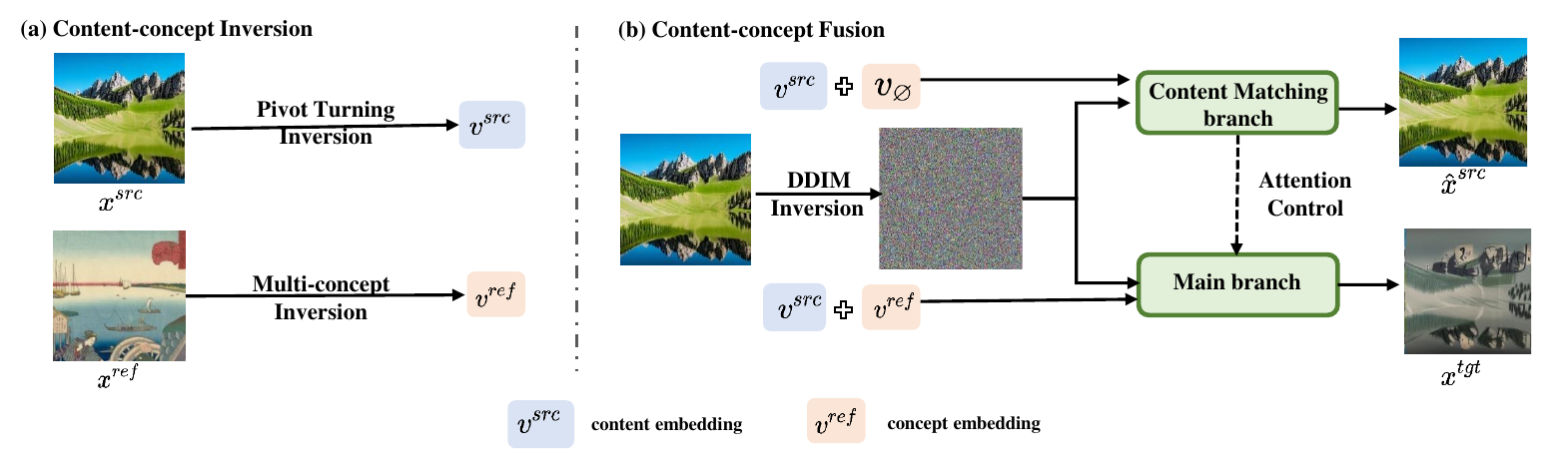}
\end{center}
   \caption{{\bf The overall visual concept translator(VCT) framework}. Given a source image $x^{src}$ and a reference image $x^{ref}$: (A) Content-concept inversion (CCI) process, we apply Pivot Turning Inversion with $x^{src}$ to obtain the source text embedding $v^{src}$. Meanwhile, we apply Multi-concept Inversion with  $x^{ref}$ to learn the reference text embedding $v^{ref}$. (B) Content-concept fusion(CCF)  process, we employ a dual-stream denoising architecture for image translation work, including a main branch $\mathcal{B}$ and a content matching branch $\mathcal{B^*}$. They start with the same initial noise inverted by $x^{src}$ using DDIM inversion. The content matching branch reconstructs the source image and extracts the attention maps for the attention control mechanism. Finally, the main branch gathers all the information to obtain a target image $x^{tgt}$.
   }
\label{fig:overall_framework}
\end{figure*}

\section{Related Works}
\subsection{Image-to-image Translation}
 The I2I aims to translate an image from the source domain to the target domain. The current I2I paradigms are mostly based on GANs \cite{baek2021rethinking, liu2021smoothing, gabbay2021scaling, zhang2020cross, zheng2021spatially, yang2022unsupervised, zhu2017unpaired}. 
 However, these methods suffer from the problem of lacking adaptability \cite{su2022dual}. The model trained in one source-target dataset cannot adapt to another one. In addition, large training images are always required for these methods.

 The TuiGAN proposed by Lin et al. \cite{lin2020tuigan} can achieve translation with only one image pair, but their method necessitates retraining the whole network for each input pair, which is very time-consuming. 

 One specific type of I2I named image style transfer tries to transform the image style from source to target.
The seminal work of Gatys et al. \cite{gatys2015neural} shows that artistic images can be generated by separating content and style with the deep neural network. Then, to realize real-time style transfer, Johnson et al. \cite{johnson2016perceptual} train a feed-forward network to handle the optimization problem mentioned by Gatys et al. Many works \cite{wu2018direction, ulyanov2016texture, ulyanov2017improved, li2016precomputed, jing2018stroke, kotovenko2019content} are categorized into per-style-per-model where the trained model can only fit one specific style. In order to increase the model flexibility, arbitrary style transfer is realized by many studies \cite{huang2017arbitrary, park2019arbitrary, jing2020dynamic, deng2020arbitrary, liu2021adaattn, sheng2018avatar, wu2020efanet} where only single forward pass is needed for any input style image. However, these methods fail to generalize to general I2I tasks such as face swap because they lack the ability to process fine-grained information.

\subsection{Diffusion-based Image Synthesis}
Large-scale diffusion models conditioned on the plain text have shown good performance in high-resolution image syntheses recently, such as Stable Diffusion \cite{rombach2022high}, Imagen \cite{saharia2022photorealistic} and DALL-E 2 \cite{ramesh2022hierarchical}. The large text-image models \cite{devlin2018bert, radford2021learning} are used by these methods to achieve text-guided synthesis.
However, the text used to generate the target images is sometimes unavailable, so the inversion technique is used by some works \cite{gal2022image, ruiz2022dreambooth, kawar2022imagic} to learn a text embedding to guide the pre-trained large-scale diffusion models. To achieve translation of images from the source to the target domain, DDIM inversion \cite{dhariwal2021diffusion, song2020denoising} finds out the deterministic noise vector with text condition along the reverse direction of the backward process, but this method is guided by text only. Our proposed method tries to handle the above drawbacks and fuses the abundant visual concepts from the image to complete the general I2I tasks.

\section{Methods}

\subsection{Preliminaries}

\noindent {\bf Latent Diffusion Models}. Diffusion models are probabilistic generative models in which an image $x_0$ is generated by progressively removing noise from an initial Gaussian noise image $x_T \sim \mathcal{N}\left(0,\mathbf{I}\right)$ in the sequence of  $x_T,x_{T-1},...,x_1,x_0$.

With the remarkable capacity of image generation, the
Latent Diffusion Model (LDM) \cite{rombach2022high} is utilized as our model backbone. Different from the conventional diffusion models that perform denoising operations directly in the image space, LDM conducts the process in the latent space with an autoencoder.

Specifically, an input image x is encoded into the latent space by the autoencoder $z=\mathcal{E}(x), \hat{x}=\mathcal{D}(z)$ (with an encoder $\mathcal{E}$ and a decoder $\mathcal{D}$) pre-trained with a large number of images. Then, the denoising process is achieved by training a neural network $\epsilon_{\theta}\left(z_t,t,v\right)$ that predicts the added noise, following the objective:
\begin{equation}
\min _\theta E_{z_0, \epsilon \sim \mathcal{N}(0, I), t \sim \operatorname{U}(1, T)}\left\|\epsilon-\varepsilon_\theta\left(z_t, t, v\right)\right\|_2^2.
\end{equation}
Note that $v$ is the text embedding generated from the text condition and $z_t$ is the noisy latent in timestamp $t$. 
$z_t$ is generated by adding noise to the sampled data $z_0$ as
\begin{equation}
z_t = \sqrt{\alpha _t} z_0 + \sqrt{1-\alpha _t} \epsilon,
\label{eq:add_noise}
\end{equation}
with $0=\alpha _t<\alpha _{t-1}<...<\alpha _0 =1$, which are hyperparameters of the diffusion schedule, and $\epsilon \sim \mathcal{N} \left(0, I\right)$.

The text embedding $v$ is obtained by $v=\tau\left(y\right)$ where $\tau$ is a BERT \cite{devlin2018bert} tokenizer and $y$ is a text prompt.
The tokenizer $\tau$ converts each word or sub-word in an input string to a token, which is an index in a specific pre-defined dictionary. Each token is then linked to a unique embedding vector that can be retrieved through an index-based lookup.

\noindent {\bf Texture inversion}. Textual Inversion (TI) \cite{gal2022image} introduces a new concept in a pre-trained text conditional generative model by learning an embedding $e^*$  as pseudo-words $S^*$. With a small collection of images $X$, TI do so by solving the following optimization problem:
\begin{equation}
\min _e E_{x~\sim\mathcal{U}_X}E_{z_t\sim q\left(z_t \mid x\right)}\left\|\epsilon-\hat\varepsilon_\theta\left(z_t, t, \tau\left(y,S^*\right)\right)\right\|_2^2.
\end{equation}
As such, it motivates the learned embedding $e^*$ to capture fine visual details unique to the concept at a coarse level.

\noindent {\bf DDIM inversion}. Inversion entails finding a noise
map $z_t$ that reconstructs the input latent code $z_0$ upon sampling. A simple inversion technique was suggested for the DDIM sampling \cite{dhariwal2021diffusion,song2020denoising}, based on the
assumption that the ODE process can be reversed in the limit of small steps:
\begin{equation}
z_{t+1}=\sqrt{\bar{\alpha}_{t+1}} f_\theta\left(z_t, t, v\right)+\sqrt{1-\bar{\alpha}_{t+1}} \varepsilon_\theta\left(z_t, t, v\right).
\label{eq:ddiminversion}
\end{equation}
where $z_{t}$ is noised latent code at timestep $t$, $\bar{\alpha}_{t+1}$ is noise scaling factor as defined in DDIM\cite{dhariwal2021diffusion}, and $f_\theta\left(z_t, t, v\right)$ predicts the final denoised latent code $z_0$.
\begin{equation}
f_\theta\left(x_t, t, c\right)=\frac{x_t-\sqrt{1-\bar{\alpha}_t} \epsilon_\theta\left(x_t, t, c\right)}{\sqrt{\bar{\alpha}_t}}
\end{equation}
In other words, the diffusion process is performed in the reverse direction, that is $z_0 \rightarrow z_T$ instead of $z_T \rightarrow z_0$, where $z_0$ is set to be the encoding of the given real image.

\noindent {\bf Classifier-free guidance}.  The diffusion model may ignore the conditional input and produce results uncorrelated with this input. One way to address this is the classifier-free guidance\cite{ho2021classifier}. During the denoising step, with a guidance scale $w \geq 1$, the classifier-free guidance prediction is defined by:
\begin{equation}
\tilde{\varepsilon}_\theta\left(z_t, t, v\right)=w \cdot \varepsilon_\theta\left(z_t, t, v\right)+(1-w) \cdot \varepsilon_\theta\left(z_t, t, v_\varnothing\right) .
\end{equation}
where $v_\varnothing$ represents the embedding of a null text.
\subsection{Overall Framework}

Given a source image $x^{src}$ and a reference image $x^{ref}$, the goal of VCT is to generate a new image $x^{tgt}$ that complies with $x^{ref}$ while preserving the structure and semantic layout of $x^{src}$. 

Fig.\,\ref{fig:overall_framework} shows the overall framework of the proposed VCT including a content-concept inversion (CCI) process and a content-concept fusion (CCF) process. As shown in Fig.\,\ref{fig:overall_framework} (a), the CCI process extracts contents and concepts from source image $x^{src}$ and reference image $x^{ref}$ into learnable embeddings. Then in Fig.\,\ref{fig:overall_framework} (b), the CCF process employs a dual-stream denoising architecture including a main branch $\mathcal{B}$ and a content matching branch $\mathcal{B^*}$, and both branches starts from the same initial noise inverted by $x^{src}$. The content matching branch reconstructs the source image and extracts the attention maps to guide the main process by the attention control mechanism. Then, the main branch gathers all information to obtain a target image $x^{tgt}$. For better understanding, we first explain the CCF process in Section\,\ref{sec:fusion}, then we describe the CCI process in Section\,\ref{sec:inversion}.

\begin{figure}[t]
\begin{center}
\includegraphics[width=\linewidth]{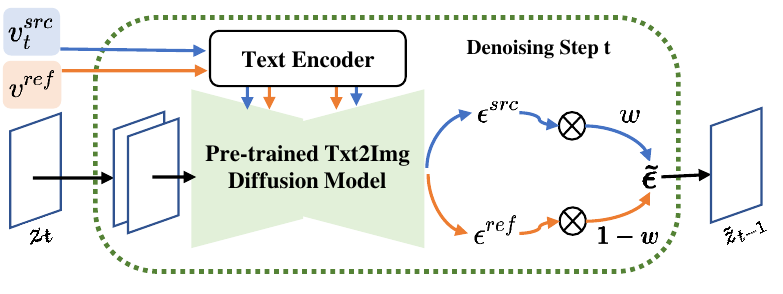}
\end{center}
\caption{ {\bf $\epsilon$ Space Fusion}. In each denoising step $t$,  the text embeddings $v_t^{src}$ and $v^{ref}$ are extrapolated with guidance scale $w$ in $\epsilon$ space.
}

\label{fig:epsilon_fusion}
\end{figure}

\begin{figure}[t]
\begin{center}
\includegraphics[width=\linewidth]{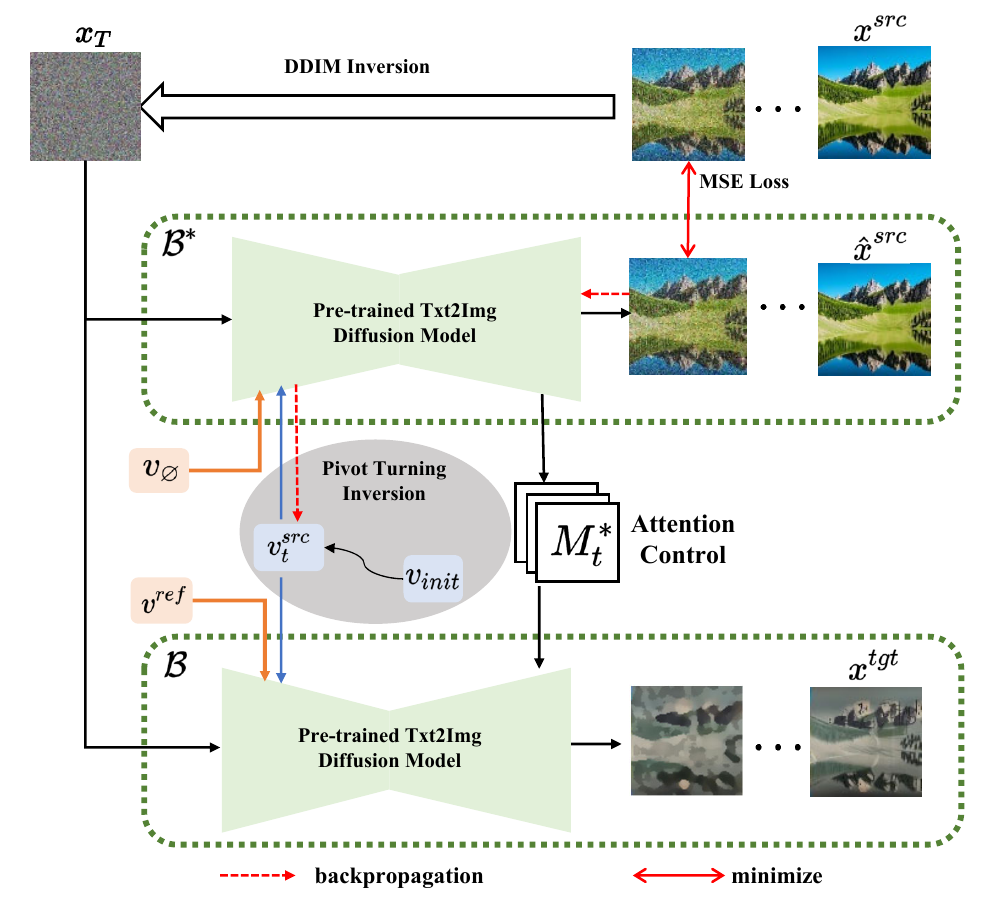}
\end{center}
   \caption{ \textbf{Dual-stream denosing architectrure.}
   }
\label{fig:content_matching_branch}
\end{figure}

\subsection{Content-concept Fusion}
\label{sec:fusion}
\noindent {\bf $\epsilon$ Space Fusion}. Given two different text embedding $v^{src}$ and  $v^{ref}$, they can be guided separately and yield two different noise prediction $\epsilon^{src}$ and $\epsilon^{ref}$:
\begin{equation}
\begin{aligned}
\epsilon^{src}=\varepsilon_\theta\left(z_t, t, v^{src}\right),\epsilon^{ref}=\varepsilon_\theta\left(z_t, t, v^{ref}\right)
.
\end{aligned}
\end{equation}
We call this space \textit{$\epsilon$ space}, as shown in Fig.\,\ref{fig:epsilon_fusion}.

According to the conclusion stated by classifier
guidance\cite{dhariwal2021diffusion} and classifier-free guidance\cite{ho2021classifier}, the noise prediction in $\epsilon$ space in each diffusion step can be interpreted as score estimation function $\varepsilon_\theta\left(z_t, t, v\right) \approx -\sigma_t \nabla_{z_t} \log p\left({z}_t\mid{v}\right)$, where $\nabla_{z_t} \log p\left({z}_t\mid{v}\right)$ represents the gradient of log-likelihood of an implicit classifier $p\left({v}\mid{z}_t\right)\propto p\left({z}_t \mid v\right) / p\left({z}_t\right).$

Under the score estimation function view of $\epsilon$ space, the independent feature  $v^{src}$ and $v^{ref}$ can be fused in the $\epsilon$ space to generate images containing certain attributes from both the source image and the reference image:
\begin{equation}
\tilde{\varepsilon}_\theta\left(z_t, t, v^{src},v^{ref}\right)=w \cdot \epsilon^{src} +(1-w) \cdot \epsilon^{ref} .
\label{eq:epsilon_fusion}
\end{equation}
$w$ is the hyperparameter that balances the two terms. It's noted that the classifier-free guidance is a special case of Eq.\,\ref{eq:epsilon_fusion}.
\\
\noindent {\bf Dual stream denoising architecture}. Based on the $\epsilon$ fusion mechanism, we now turn to the image translation task. As shown in Fig.\,\ref{fig:content_matching_branch}, let $x^T$ be the initial noise, obtained by inverting $x^{src}$ using DDIM inversion with Eq.\,\ref{eq:ddiminversion}, where we set $v=v_\varnothing$.
Starting from the same initial noise $x^T$, we employ a dual-stream denoising architecture for I2I, denoted as a main branch $\mathcal{B}$ and a content matching branch $\mathcal{B^*}$.

The content matching branch $\mathcal{B^*}$ is a denoising process that perfectly reconstructs the source image $x^{src}$ (with $z^{src}$ perfectly reconstructed in latent space for LDMs), and the main branch $\mathcal{B}$ is the denoising process that finally serves for the I2I tasks. 
\begin{equation}
\begin{aligned}
\mathcal{B^*}&:z_T \rightarrow z_{T-1}^* \rightarrow... \rightarrow z_1^* \rightarrow z^{src} \\
\mathcal{B}&:z_T \rightarrow z_{T-1} \rightarrow... \rightarrow z_1\rightarrow z^{tgt} .
\end{aligned}
\end{equation}

At each denoising step $t$, the content matching branch $\mathcal{B^*}$ aims to extract the text embedding $v_t^{src}$ and the attention map $M_t^*$, which would serve for the parallel denoising step in the main branch. With $\mathcal{B^*}$,  we obtain meaningful embedding and generated structure of the source image.

To better inject the information of the source image $x^{src}$, the dual stream diffusion processes have almost the same computation pipelines, except for the reference embeddings used in $\epsilon$ space fusion. We perform  $\epsilon$ space fusion in the content matching branch as the main branch by:
\begin{equation}
\tilde{\varepsilon}_\theta\left(z_t, t, v^{src},v_\varnothing\right)=w \cdot \epsilon^{src} +(1-w) \cdot \epsilon^{\varnothing} .
\label{eq:cfg_fusion}
\end{equation}
The above sampling procedure reduces to the classifier-free guidance. And we should ensure that Eq.\,\ref{eq:epsilon_fusion} and Eq.\,\ref{eq:cfg_fusion} have the same $w$ for the dual-stream diffusion architecture.

\noindent {\bf Attention control}. Recent large-scale diffusion models \cite{rombach2022high,saharia2022photorealistic,ramesh2022hierarchical} incorporate conditioning by augmenting the denoising network $\varepsilon_{\theta}$ with self-attention layer and cross-attention layer \cite{bahdanau2014neural,vaswani2017attention}. Of particular interest are the \textit{cross-attention map} and \textit{self-attention map}, denoted as $M$ in total, which is observed to have a tight relation with the structure of the image \cite{hertz2022prompt}. To this end, Amir at al.\cite{hertz2022prompt} pursue \textit{prompt-to-prompt} editing framework for text-guided image translation task, which controls the attention maps of the edited image by injecting the attention maps of the original image along the diffusion process. 

In our case, we employ soft attention control as described in \textit{prompt-to-prompt} \cite{hertz2022prompt}. Let $M_t^*$ be the attention map of a single step $t$ of the content matching branch, and $M_t$ be the attention map of the main branch. The soft attention control is defined as:
\begin{equation}
\widehat{M} = \textit{AC}\left(M_t,M_t^*,t\right) = \begin{cases}M_t^* & \text { if } t<\tau \\ M_t & \text { otherwise }\end{cases}
\end{equation}
where $\tau$ is a timestamp parameter that determines until which step the attention map replacement is applied. We define $\tilde{\varepsilon}_\theta\left(z_t, t, v^{src},v^{ref}\right)\{M \leftarrow  \widehat{M}\}$ to be the function that overrides the attention map $M$ in $\tilde{\varepsilon}$ with additional given map $\widehat{M}$.

\subsection{Content-concept Inversion}
\label{sec:inversion}

\noindent {\bf Pivotal turning inversion} is proposed to generate the content embedding to guide the CCF process.  We start by studying the DDIM inversion\cite{dhariwal2021diffusion,song2020denoising}. In practice , a slight error is incorporated in every step. For unconditional diffusion models, the accumulated error is negligible and the DDIM inversion succeeds. However, recall that meaningful editing using the Stable Diffusion model \cite{rombach2022high} requires applying classifier-free guidance with a guidance scale $w$. Ron et al.\cite{mokady2022null} have presented that such a guidance scale amplifies the accumulated error. 

To this end, Ron et al. \cite{mokady2022null} introduce null-text inversion technology to reconstruct the image and further for text-guided image translation tasks. Null-text inversion modifies the unconditional embedding in each timestamp $t$ that is used for classifier-free guidance to match the initial conditional DDIM inversion trajectory. 

In our image-guided case, we do not know the exact text prompt of the source image $x^{src}$. So, inspired by \cite{mokady2022null}, we implement unconditional DDIM inversion, and optimize the source embedding $v_t^{src}$ in each timestamp $t$ for accurately matching the source image $x^{src}$, instead of the DDIM inversion trajectory.

In each timestamp $t$, we optimize the $v_t^{src}$ by:
\begin{equation}
\min _{v_t^{src}} \left \| z_0- \hat{z}_0\left(z_t,v_t^{src}\right) \right\|^2_2
\end{equation}
where $\hat{z}_0\left(z_t,v_t^{src}\right)$ refers to the estimated clean latent $\hat{z}_0$ given $z_t$ and $v_t^{src}$, using the Tweedie’s formula \cite{kim2021noise2score}. We rewrite it as:
\begin{equation}
\hat{z}_0\left(z_t,v_t^{src}\right)=\frac{z_t}{\sqrt{\bar{\alpha}_t}}-\frac{\sqrt{1-\bar{\alpha}_t}}{\sqrt{\bar{\alpha}_t}} \tilde{\varepsilon}_\theta\left(z_t, t, v^{src},v_\varnothing\right)
\label{eq:Tweedie}
\end{equation}
where $\tilde{\varepsilon}_\theta\left(z_t, t, v^{src},v_\varnothing\right)$ is defined in Eq.\,\ref{eq:cfg_fusion}. 

Note that for every $t<T$ , the optimization should start from the endpoint of the previous step $t + 1$ optimization, which computes a constant $z_t$ using the optimized $v_{t+1}^{src}$ and $z_{t+1}$. Otherwise, the learned embedding would not hold at inference.

\noindent {\bf Multi-concept inversion} is proposed to represent complex visual concepts by generating the concept embedding. Lastly, we should learn a reference embedding $v^{ref}$ from the reference image $x^{ref}$. The methodological approach is related to Textual Inversion \cite{gal2022image} and DreamArtist \cite{dong2022dreamartist}. 

To represent the concepts in the input images, Textual Inversion \cite{gal2022image} learns an embedding as pseudo-words $S_*$ from few-shot images. DreamArtist \cite{dong2022dreamartist} improves Textual Inversion, which learns a paired positive and negative multi-concept embeddings($S_*^p$ and $S_*^n$) from one-shot image and proposes reconstruction constraint for detail enhancement. In our case, we apply a similar strategy as DreamArtist, yet our method offers two improvements:

Firstly, we find that the multi-concept embeddings are useful for mining semantics information from the images, while the negative embeddings are optional. And in our pipeline, the negative embeddings are in conflict with the source embedding $x^{src}$. Thus, we use single positive multi-concept embeddings for learning the reference text embedding $v^{ref}$. We freeze the parameters of the generative diffusion model $\varepsilon_\theta$, and optimize the $v^{ref}$ using the denoising diffusion objective\cite{ho2020denoising}:
\begin{equation}
\mathcal{L}_{ldm}= E_{\epsilon,t } \left[\left\|\epsilon-\varepsilon_\theta\left(z_t^{ref}, t, v^{ref}\right)\right\|_2^2\right].
\end{equation}
where $v^{ref}$ is the multi-concept embeddings, $z_t^{ref}$ is a noisy version of $z^{ref}_0$ (the latent code of the reference image $x^{ref}$) obtained using Eq.\,\ref{eq:add_noise}, $\epsilon \sim \mathcal{N}(0, I)$ and $t \sim \operatorname{U}(1, T) $.

Secondly, we improve the reconstruction constraint for the mechanism of detail enhancement in DreamArtist. DreamArtist applys reconstruction constraint in $x$ space, which can be denoted as $\mathcal{D}\left(\hat{z}_{t-1}\left(z_t,S^*\right)\right) \leftrightarrow x_0$.
For one thing, optimization in $x$ space suffers from huge resource consumption due to the gradient backpropagation inside decoder $\mathcal{D}$. For another thing, there is a gap between the estimated $z_{t-1}$ and $z_0$, especially in the early stage of the denoising process.

Formally, we implement reconstruction constraint in $z$ space. The reconstruction loss can be written as:
\begin{equation}
\mathcal{L}_{rec}= E_{\epsilon,t } \left[\left\|z_0^{ref}-\hat{z}_0\left(z_t^{ref},v_t^{ref}\right)\right\|_2^2\right].
\end{equation}
where $\hat{z}_0\left(z_t^{ref},v_t^{ref}\right)$ refers to the estimated clean latent $\hat{z}_0^{ref}$ given $z_t^{ref}$ and $v_t^{ref}$, using Eq.\,\ref{eq:Tweedie}.

\section{Experiments}

\begin{figure*}[htb]
\begin{center}
\includegraphics[width=\linewidth]{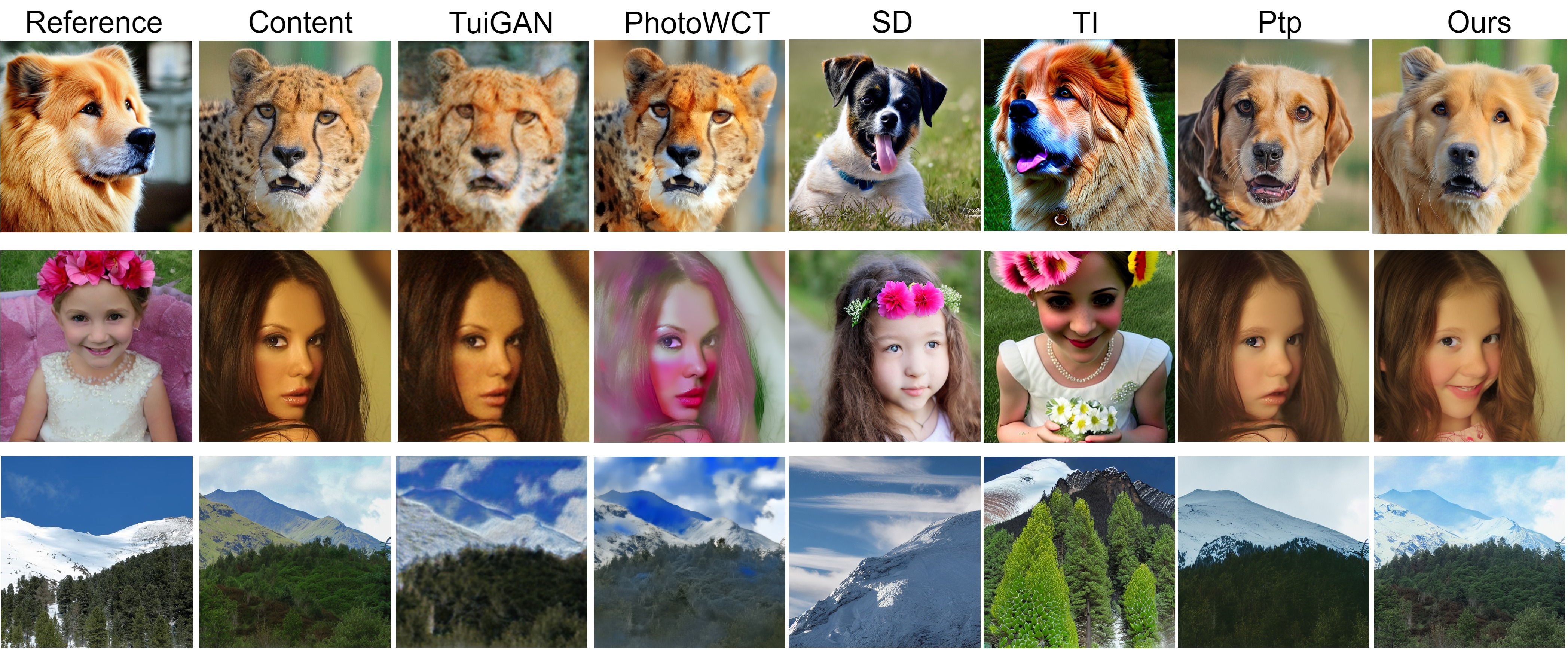}
\end{center}
   \caption{ Model Performance in general image-to-image translation task including leopard$\rightarrow$dog, face swap and mountain$\rightarrow$snow mountain. The first and second columns show the reference and content images, respectively. The 3-7 columns show the translation results of different methods.
   }
\label{fig:general_I2I}
\end{figure*}

\begin{figure}[t]
\begin{center}
\includegraphics[width=\linewidth]{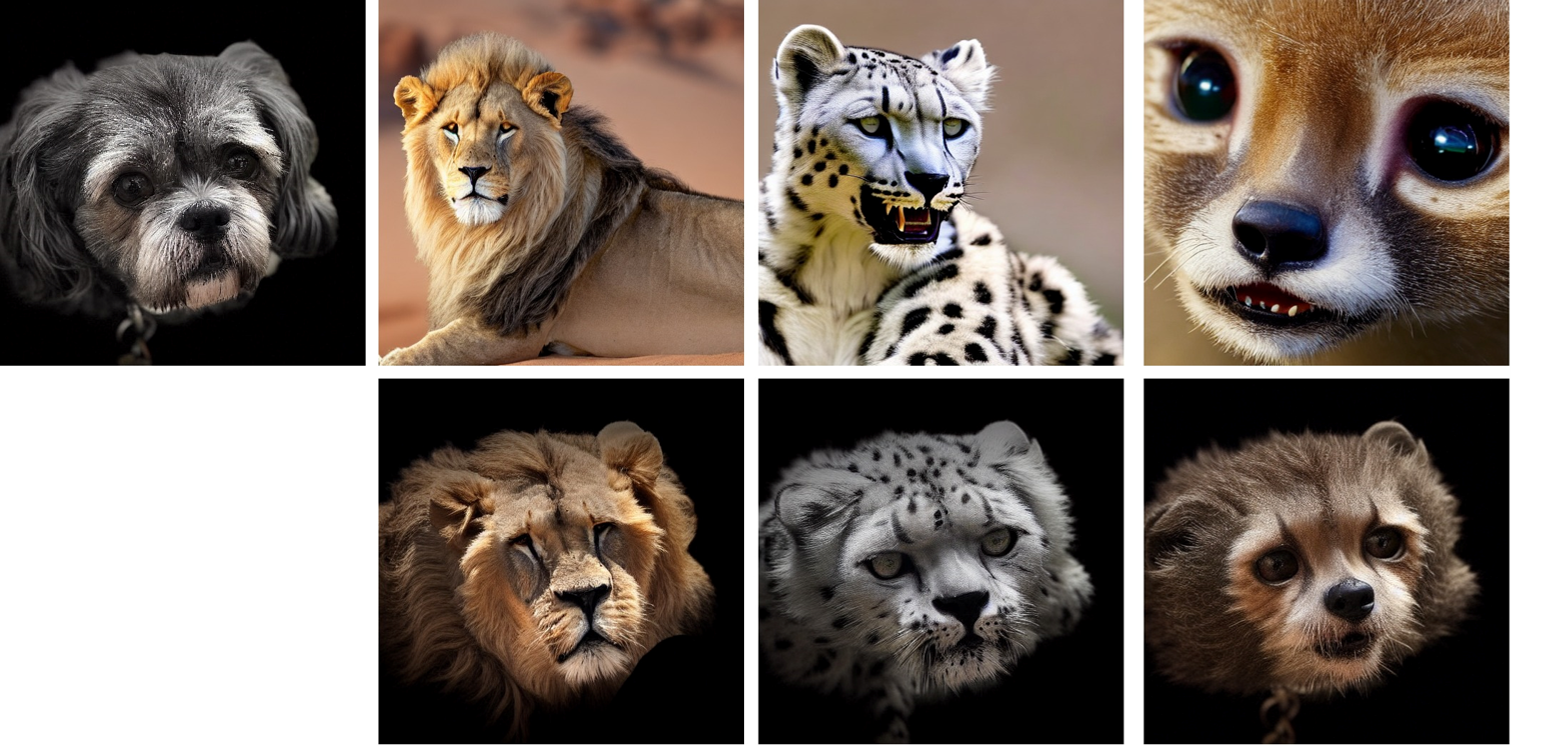}
\end{center}
   \caption{ Our method can perform fine-grained image-to-image translation. 
   }
\label{fig:I2I_fine_grade}
\end{figure}

\begin{figure}[t]
\begin{center}
\includegraphics[width=\linewidth]{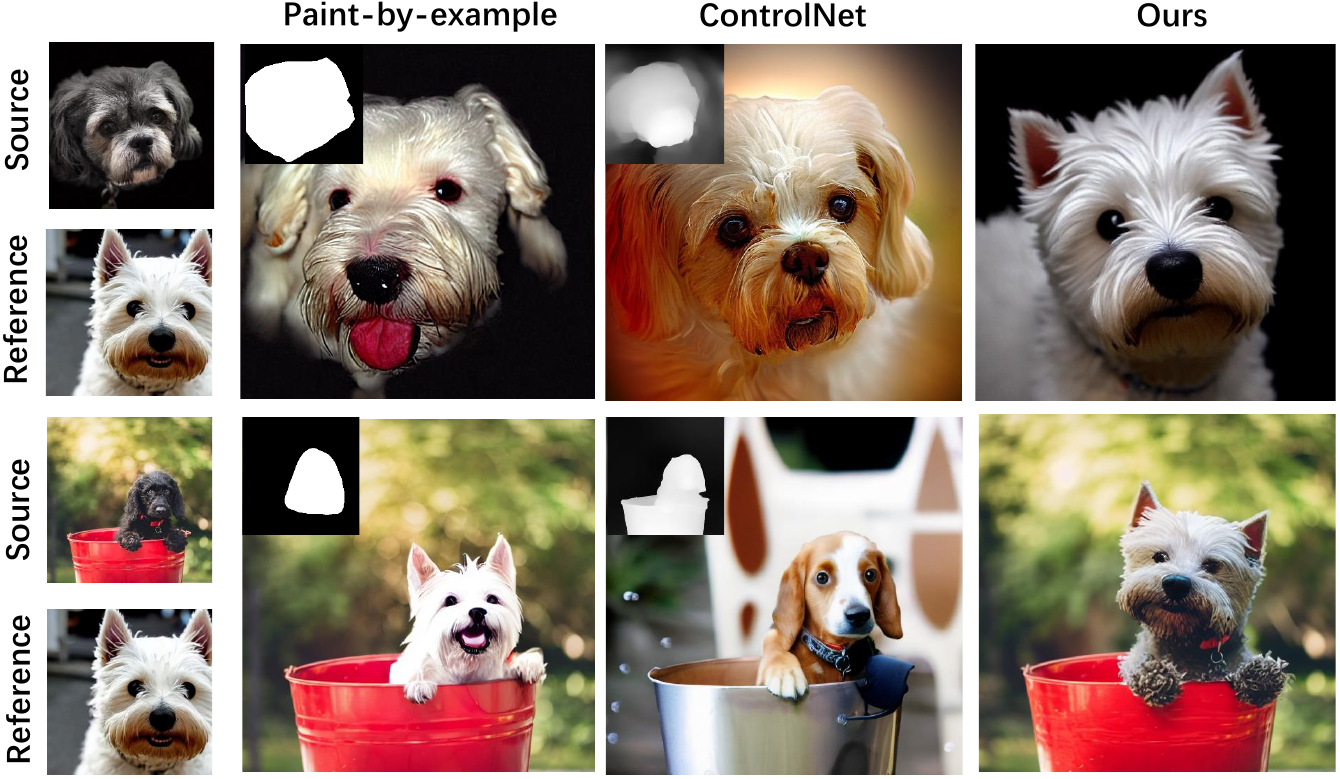}
\end{center}
   \caption{ Comparisons to concurrent one-shot baselines: Paint-by-example\cite{yang2022paint} and ControlNet\cite{zhang2023adding}. 
   }
\label{fig:concurrent}
\end{figure}

\begin{figure*}[t]
\begin{center}
\includegraphics[width=\linewidth]{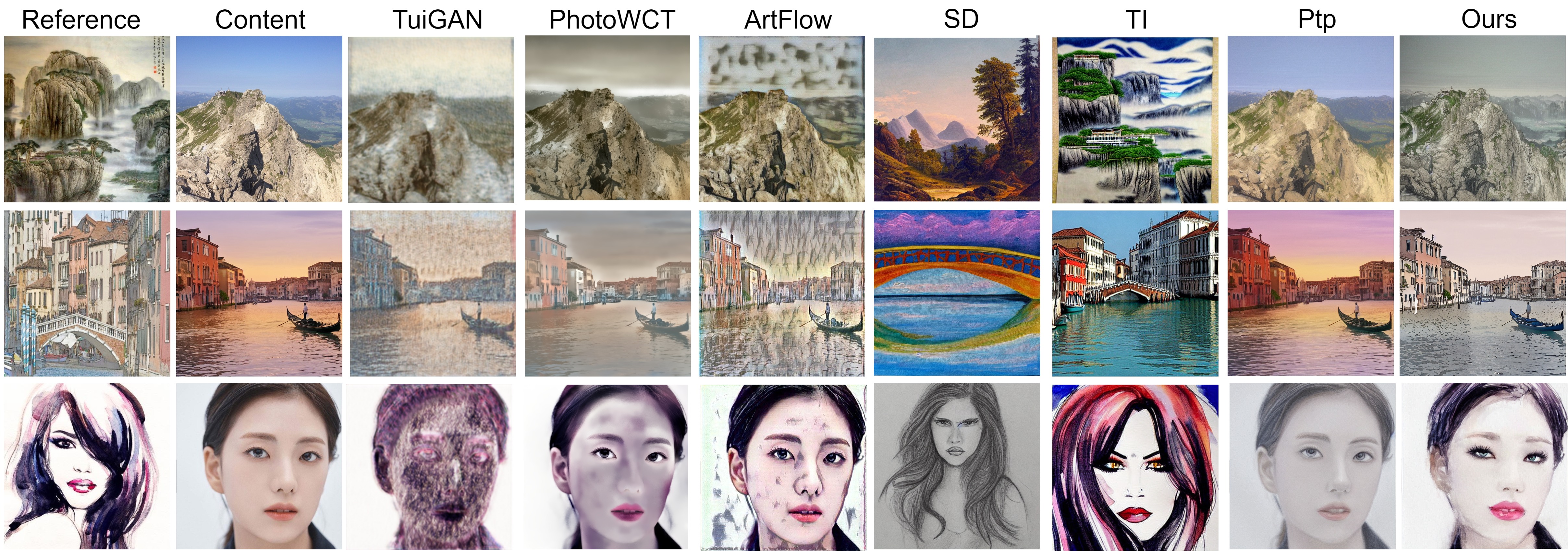}
\end{center}
   \caption{ Model Performance in the style translation task. The first and second columns show the reference and content images, respectively. The 3-7 columns show the translation results of different methods.
   }
\label{fig:style_transfer}
\end{figure*}

\begin{figure}[t]
\begin{center}
\includegraphics[width=\linewidth]{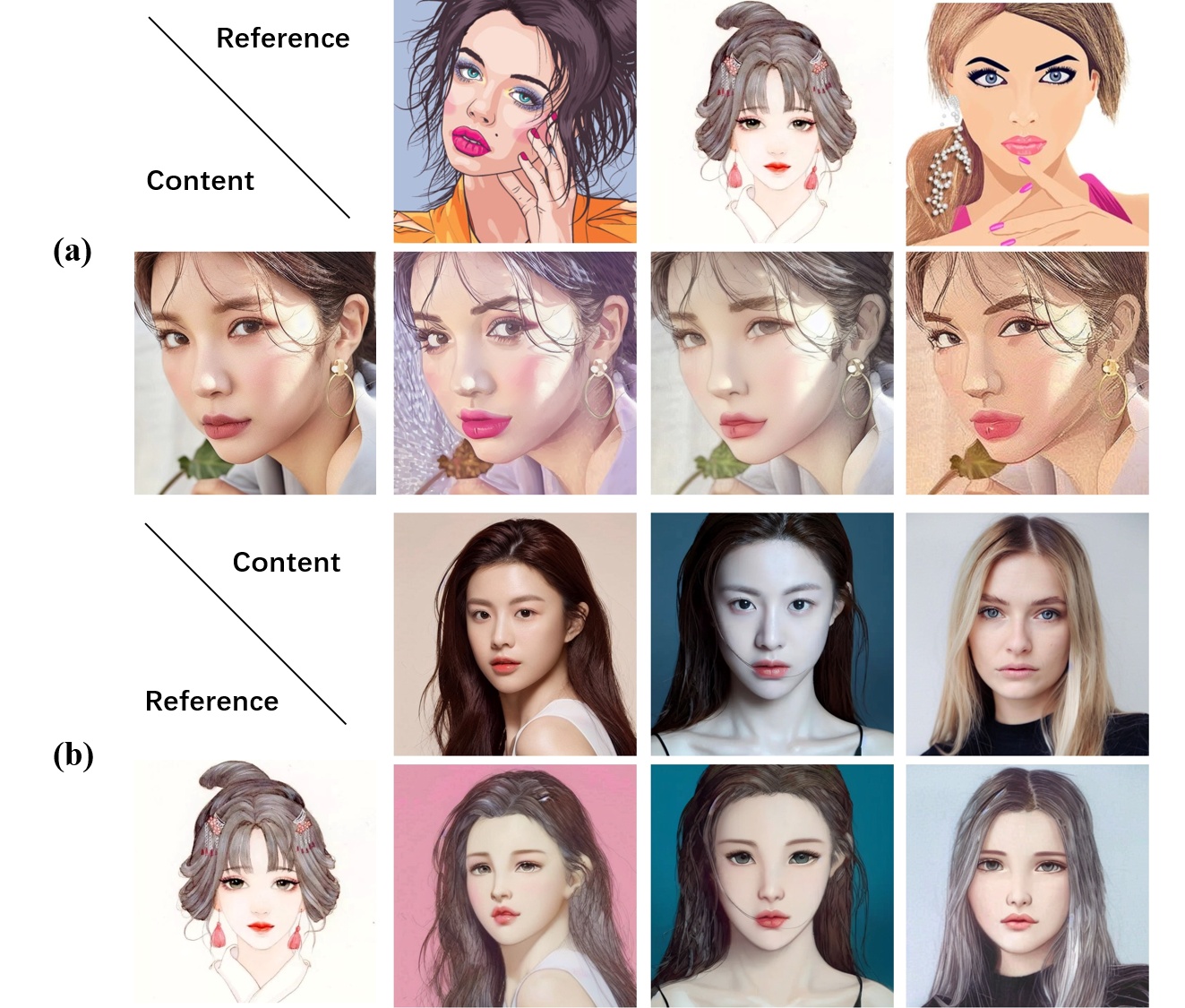}
\end{center}
   \caption{ Content variance and style variance of the style transfer result.
   }
\label{fig:style_transfer_fine_grade}
\end{figure}

\subsection{Implementation details}
Putting all components together, our full algorithm is presented in our supplementary material. The core training process consists of two parts: pivotal tuning inversion with $x^{src}$ and multi-concept inversion with $x^{ref}$, which can be implemented independently. For more details please refer to our supplementary material.

Our experiments were conducted using a single A100 GPU. We use Adam\cite{Kingma2014AdamAM} optimizer for both training processes. We collect the evaluation images from the large-scale LAION 5B dataset \cite{schuhmann2021laion} containing 5 billion images.

\subsection{Comparison to Prior/Concurrent Work}

\noindent {\bf General I2I Tasks}. Here, we evaluate the performance of the proposed framework in general I2I tasks including leopard$\rightarrow$dog, face swap, and mountain$\rightarrow$snow mountain, as shown in Fig.\,\ref{fig:general_I2I}. We compare the proposed method with TuiGAN \cite{lin2020tuigan}, PhotoWCT \cite{li2018closed}, stable diffusion (SD) \cite{rombach2022high}, textual inversion (TI) \cite{gal2022image} and prompt-to-prompt (Ptp) \cite{hertz2022prompt}. For text-to-image models without learned embedding input including SD and Ptp, we use BLIP image caption model \cite{li2022blip} to extract text description as input of diffusion model.
From Fig.\,\ref{fig:general_I2I}, the GAN-based translation methods TuiGAN and PhotoWCT cannot well translate the concept with only one image input with poor generation quality. 
For example, from columns 3-4 of Fig.\,\ref{fig:general_I2I}, GAN-based methods only translate part of texture features from the reference image in leopard$\rightarrow$dog and face swap task, and the image quality is poor in the mountain$\rightarrow$snow mountain task. Therefore, the GAN-based methods cannot achieve satisfactory results in the one-shot setting.
For diffusion-based methods SD and TI, the concepts of the reference image can be well preserved, but the information in the content image cannot be extracted. As shown in column 7 of Fig.\,\ref{fig:general_I2I}, Ptp can well preserve content but the concepts in the reference images cannot be fused. By tackling all weaknesses of the above methods, the proposed VCT can generate the best results with concepts learned and content preserved.

Furthermore, to evaluate the strong concept translation ability of the proposed VCT, we keep the content image fixed and change different reference images in Fig.\,\ref{fig:I2I_fine_grade}. The generation results of different reference images show satisfactory content preservation and concept translation ability. More results can be found in the supplementary material.

As shown in Fig.\,\ref{fig:concurrent}, we further make comparisons to concurrent one-shot baselines: Paint-by-example\cite{yang2022paint} and ControlNet\cite{zhang2023adding}. These methods use additional conditions for controlling the generated image, while our method obtains better performance.

\begin{figure}[h]
\begin{center}
\includegraphics[width=\linewidth]{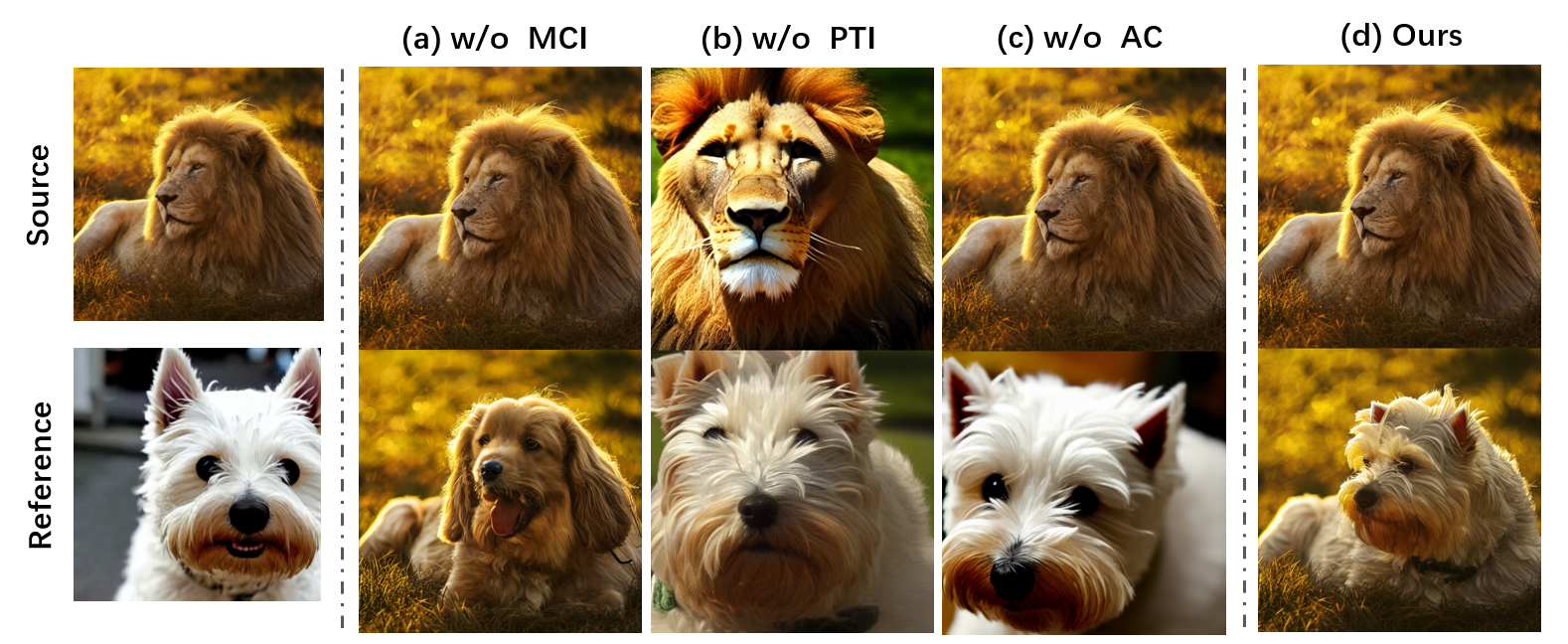}

\end{center}

 \vspace{-0.25cm}
   \caption{ Visualization of the ablation study. 
   }
\label{fig:ablation_study_method}
\end{figure}

\noindent {\bf Image Style Transfer}. In addition to general I2I, the proposed method also achieves excellent results in tasks of image style transfer. We compare our method with recent SOTAs in style transfer tasks with different art styles. As shown in Fig.\,\ref{fig:style_transfer}, we totally compare with three GAN-based methods including TuiGAN \cite{lin2020tuigan}, PhotoWCT \cite{li2018closed} and ArtFlow \cite{an2021artflow}, and three diffusion-based methods including SD \cite{rombach2022high}, TI \cite{gal2022image} and Ptp \cite{hertz2022prompt}. Following the setting of general I2I, we use the BLIP image caption model to extract text descriptions for text-to-image model SD and Ptp. From the results in Fig.\,\ref{fig:style_transfer}, large defects exist for results generated by GAN-based methods, especially for TuiGAN and ArtFlow as columns 3 and 5 in Fig.\,\ref{fig:style_transfer}. The same content preservation problem exists in diffusion-based methods SD and TI as general I2I. For Ptp, although the contents are preserved, the concept in the reference images cannot be well translated. The proposed method can also generate the most satisfactory images, as shown in column 9 of Fig.\,\ref{fig:style_transfer}.

We also evaluate the model performance by keeping the reference image fixed and changing the content image, and vice versa. The results are shown in Fig.\,\ref{fig:style_transfer_fine_grade}. The excellent translation results prove the generalization of the proposed method.


\begin{table}[htbp]
\small
\setlength{\belowcaptionskip}{-0.0cm}
\centering
\caption{Quantitative evaluation results.}
\label{tab:quan}
\begin{tabular}{|m{2cm}<{\centering}|m{2cm}<{\centering}|p{2cm}<{\centering}|}
\hline
& LPIPS$\downarrow$  & CLIPscore$\uparrow$ \\
\hline
Artflow\cite{an2021artflow} & 0.42 & 0.51\\
\hline
CAST\cite{zhang2022domain} & 0.55 & 0.61 \\
\hline
InST\cite{zhang2023inversion} & 0.43 & 0.48 \\
\hline
StyleFormer\cite{wu2021styleformer} & 0.46 & 0.48 \\
\hline
StyTR2 \cite{deng2022stytr2}  & 0.43 & 0.53 \\
\hline
Ours & \textbf{0.35} & \textbf{0.66}\\
\hline
\end{tabular}
\end{table}





 \noindent {\bf{Quantitative Comparision}}. Due to the absence of ground truth for style transfer and the domain gap between the two domains, quantitative evaluation remains an open challenge. 
Following the same setting of StyTR2, we randomly choose 800 generated images from different translation tasks for quantitative comparison. We compare the proposed method with the state-of-the-art including Artflow \cite{an2021artflow}, CAST \cite{zhang2022domain}, InST \cite{zhang2023inversion}, StyleFormer \cite{wu2021styleformer} and StyTR2 \cite{deng2022stytr2}, and the results are shown in Table\,\ref{tab:quan}. We use Learned Perceptual Image Patch Similarity (LPIPS) to evaluate the difference between output and source image, and CLIP score to evaluate the difference between output and reference image. The results show our proposed method can achieve the best performance with the lowest LPIPS and highest CLIPscore.


\subsection{Ablation Study}

Finally, we ablate each component of our method and show its effectiveness, including multi-concept inversion (MCI), pivotal turning inversion (PTI), and attention control (AC).  

See visual ablation studies in Fig.\,\ref{fig:ablation_study_method}:  (a) By removing MCI, where we use the word 'dog' to generate the reference embedding $v^{ref}$ in our pipeline, the generated result is not the specific dog in the reference image. (b) Without using PTI, the content matching branch cannot reconstruct the content image, due to the inconsistent DDIM sampling trajectory. (c) By removing AC, the result can not retain the structure of the content image. 

Overall, we can obtain the best generation outputs by using all of our proposed components, which better preserves the structure and semantic layout of the content image, while complying with the reference image. Further ablations can be found in the supplementary material.

\section{Conclusion}
In this work, motivated by the importance of visual concepts in our daily life,
we complete the general I2I with image guidance by proposing a novel framework named VCT. It can preserve the content in the source image and translate visual concepts guided by a single reference image. We evaluate the proposed model on a wide range of general image-to-image translation tasks with excellent results.

\raggedbottom
\pagebreak
\clearpage

\appendix

\noindent{\Large \textbf{{Appendix}}}

\section{More Implentation Details}
\begin{algorithm}[h] 
	\caption{Visual Concept Translator} 
	\label{alg::conjugateGradient}
	\begin{algorithmic}[1] 
		\Require 
		$x^{src}$, $x^{ref}$: source and reference image.
            \Require 
            $\alpha$, $\beta$: learning rates; $S_{m}$, $S_{p}$: training steps.
		\State \textbf{Initial} content embedding $v^{src}$ and concept embedding $v^{ref}$.
            \State $\triangleright$ Multi-concept inversion
            \For{step=1,...,$S_{m}$}
            \State Compute $L_{ldm}$ and $L_{rec}$ in Eq.\;(14) and Eq.\;(15);
            \State Update concept embedding with gradient descent: $v^{ref}\longleftarrow v^{ref}-\alpha\bigtriangledown_{v^{ref}}(L_{ldm}+L_{rec})$.
            \EndFor
            \State $\triangleright$ Pivotal turning inversion
            \State Compute source latent $z^{src}=\mathcal{E}(x^{src})$ with encoder $\mathcal{E}$;
            \State Compute $z_{T}=$DDIM-Inversion$(z^{src})$ with Eq.\;(4)
            \State Compute unconditional embedding $v_\varnothing = \tau (\varnothing)$ with tokenizer $\tau$;
            \For{t=T,...,1};
                \For{step=1,...,$S_{p}$}
                \State Compute $L=z_0- \hat{z}_0\left(z_t,v_t^{src}\right)$ in Eq.\;(12);
                \State Update content embedding with gradient descent: $v_t^{src}\longleftarrow v_t^{src}-\alpha\bigtriangledown_{v_t^{src}}L$
                \EndFor
                \State $\hat{\epsilon} \leftarrow \tilde{\varepsilon}_\theta\left(z_t, t, v^{src}_t,v^{\varnothing}\right)$ in Eq.\;(10);
                \State $z_{t-1}\leftarrow \text{DDIM-sample}(z_t,\hat{\epsilon},t)$
            \EndFor
            \State $\triangleright$ Content-concept fusion
            
            \For{t=T,...,1};
            \State Compute noise prediction $\epsilon^{src}$ and  $\epsilon^{ref}$ in Eq.\;(7);
             \State $M_t,\hat{\epsilon}\leftarrow \tilde{\varepsilon}_\theta\left(z_t, t, v^{src}_t,v^{ref}\right)$ in Eq.\;(8);
             \State  $M_t^*,\hat{\epsilon}^*\leftarrow \tilde{\varepsilon}_\theta\left(z^{*}_t, t, v^{src}_t,v^{\varnothing}\right)$ in Eq.\;(10);

            \State $\widehat{M}_t \leftarrow \textit{AC}\left(M_t,M_t^*,t\right)$ in Eq.\;(11)
            \State $\hat{\epsilon}=\tilde{\varepsilon}_\theta\left(z_t, t, v^{src}_t,v^{ref}\right)\{M \leftarrow  \widehat{M}_t\}$
            \State $z_{t-1}\leftarrow \text{DDIM-sample}(z_t,\hat{\epsilon},t)$
            \State $z^*_{t-1}\leftarrow \text{DDIM-sample}(z^*_t,\hat{\epsilon}^*,t)$
            \EndFor
        \State Compute target image $x^{tgt}=\mathcal{D}(z^{0})$ with decoder $\mathcal{D}$; 
    
	\end{algorithmic} 
 \label{algorithm}
\end{algorithm}

Our full algorithm is shown in Algorithm \ref{algorithm}. For multi-concept inversion, we empirically found that 200 training steps are enough for convergence, and this process only takes about 150 seconds. Furthermore, for 
pivotal turning inversion, our found optimal training step is 1000, which takes about 60 seconds. The learning rate is $5\times10^{-4}$ for multi-concept inversion. For pivotal turning inversion, we reduce the learning rate when step increases, as
\begin{equation}
\begin{aligned}
lr = 1\times 10^{-2} \times s / 5000,
\end{aligned}
\end{equation}
where $s$ is the current step numbers.
The algorithm also includes the unconditional embedding $v_\varnothing$, which is extracted by putting empty text to the BERT tokenizer. The Adam optimizer is used for both inversion processes.

\section{More Results of General Image-to-image Translation}
\begin{figure*}[t]
\begin{center}
\includegraphics[width=\linewidth]{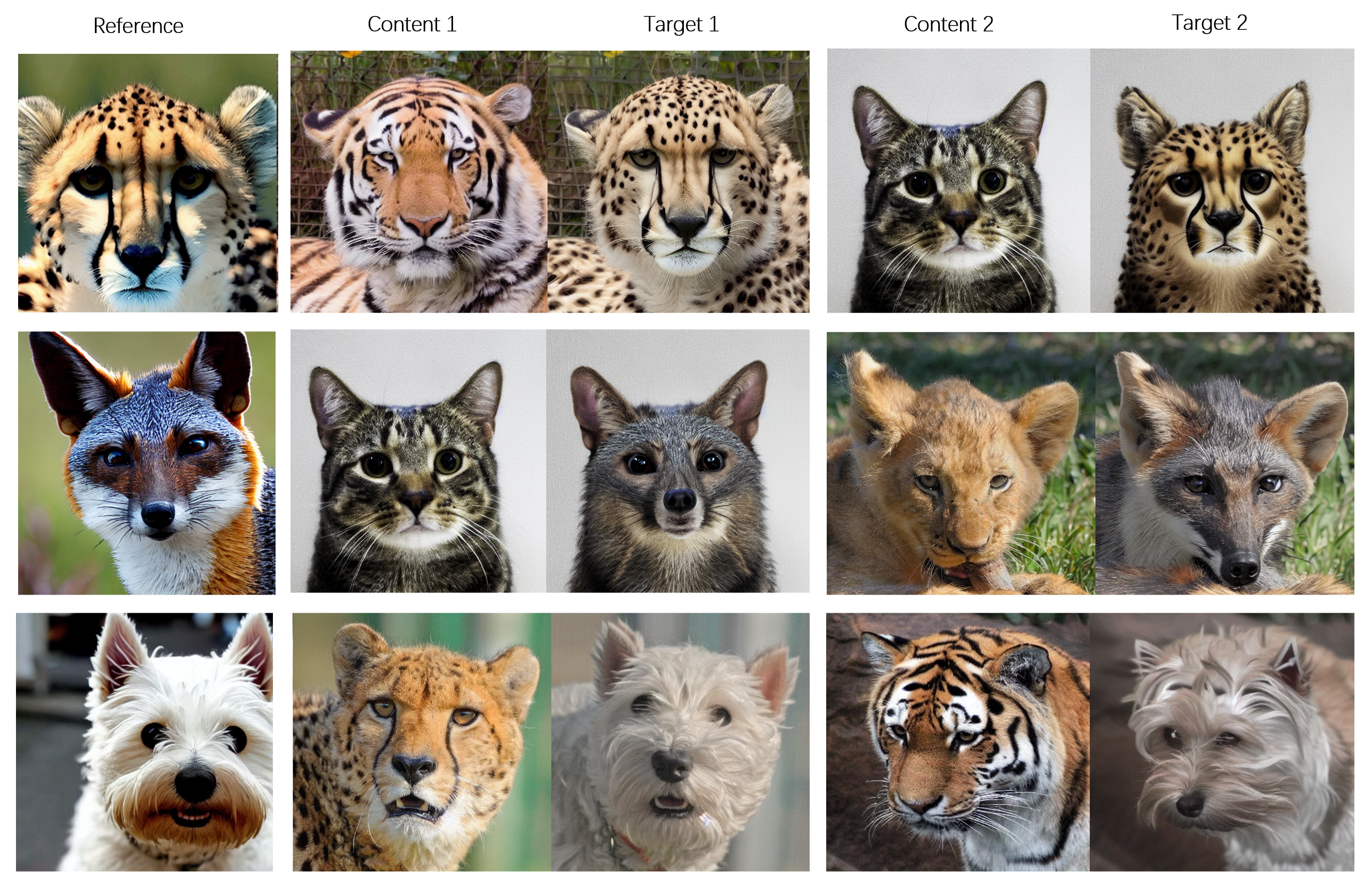}
\end{center}
\caption{ Model performance in general image-to-image translation tasks. The first column contains the reference images, and the following columns contain two groups of results based on the content images in column 2 and column 4. Our model generates realistic samples that reflect the reference image while maintaining the structure of the source image.
}
\label{appendixfig:appendix_animal_face}
\end{figure*}

To further verify the model performance in the general image-to-image translation tasks, we make more experiments with different reference images, as shown in Fig.\;\ref{appendixfig:appendix_animal_face}. 

\begin{figure*}[t]
\begin{center}
\includegraphics[width=\linewidth]{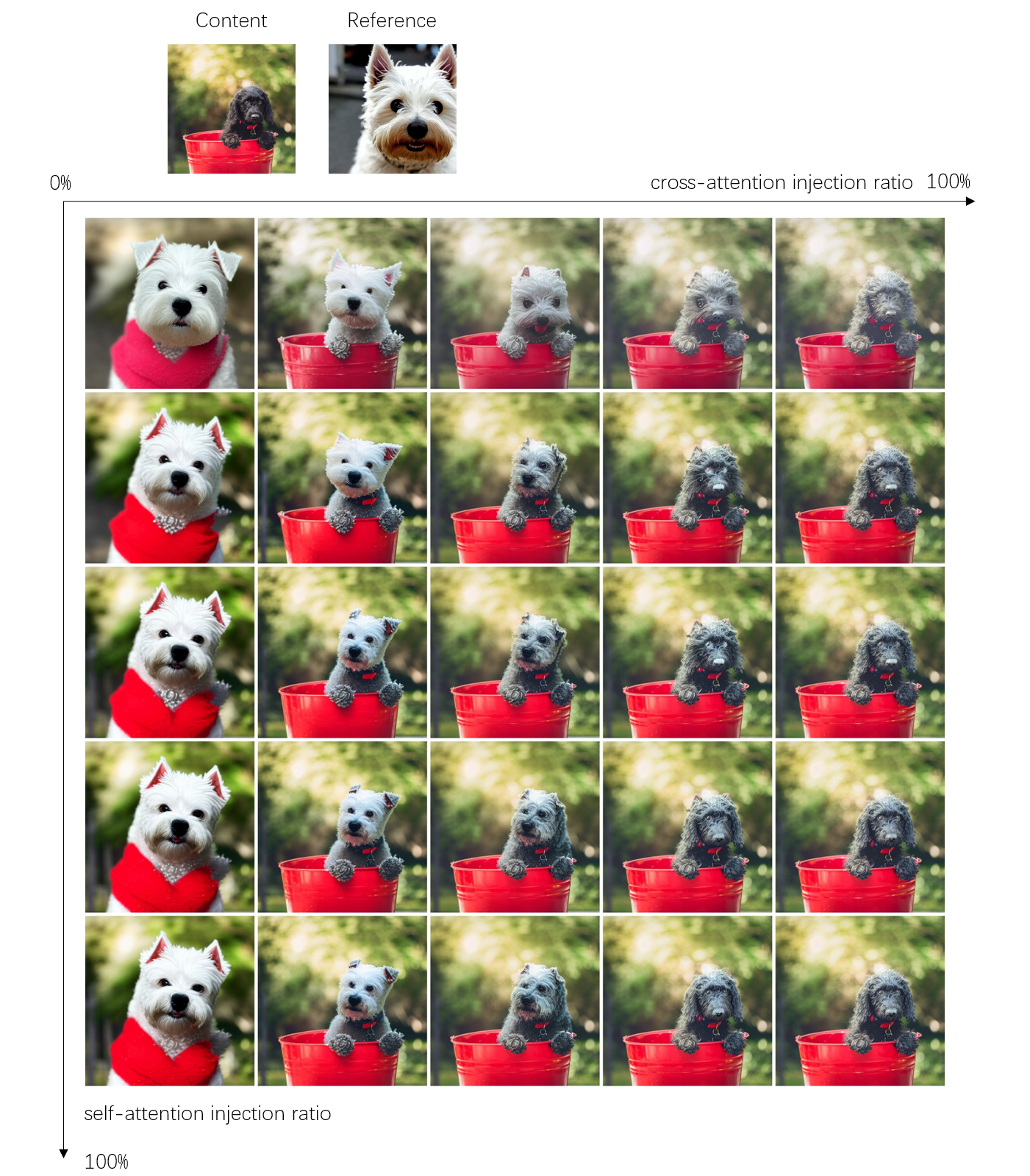}
\end{center}
\caption{ Trade-off between structural preservation and semantic changes. We generate the ideal results by adjusting the cross-attention and self-attention injection ratios to optimal values. 
}
\label{appendixfig:result_balance}
\end{figure*}

It's noted that there is a trade-off between structural preservation and semantic changes. As shown in Fig.\;\ref{appendixfig:result_balance}, the injection ratio of cross-attention and self-attention affects the result a lot. To obtain the ideal results, we can adjust the attention injection ratio to the optimal value. Empirically, we adopt a low cross-attention injection ratio of about $20\%$, and adjust the self-attention injection ratio to achieve different preservation results.

\section{More Results of Style Transfer}
\begin{figure*}[t]
\begin{center}
\includegraphics[width=\linewidth]{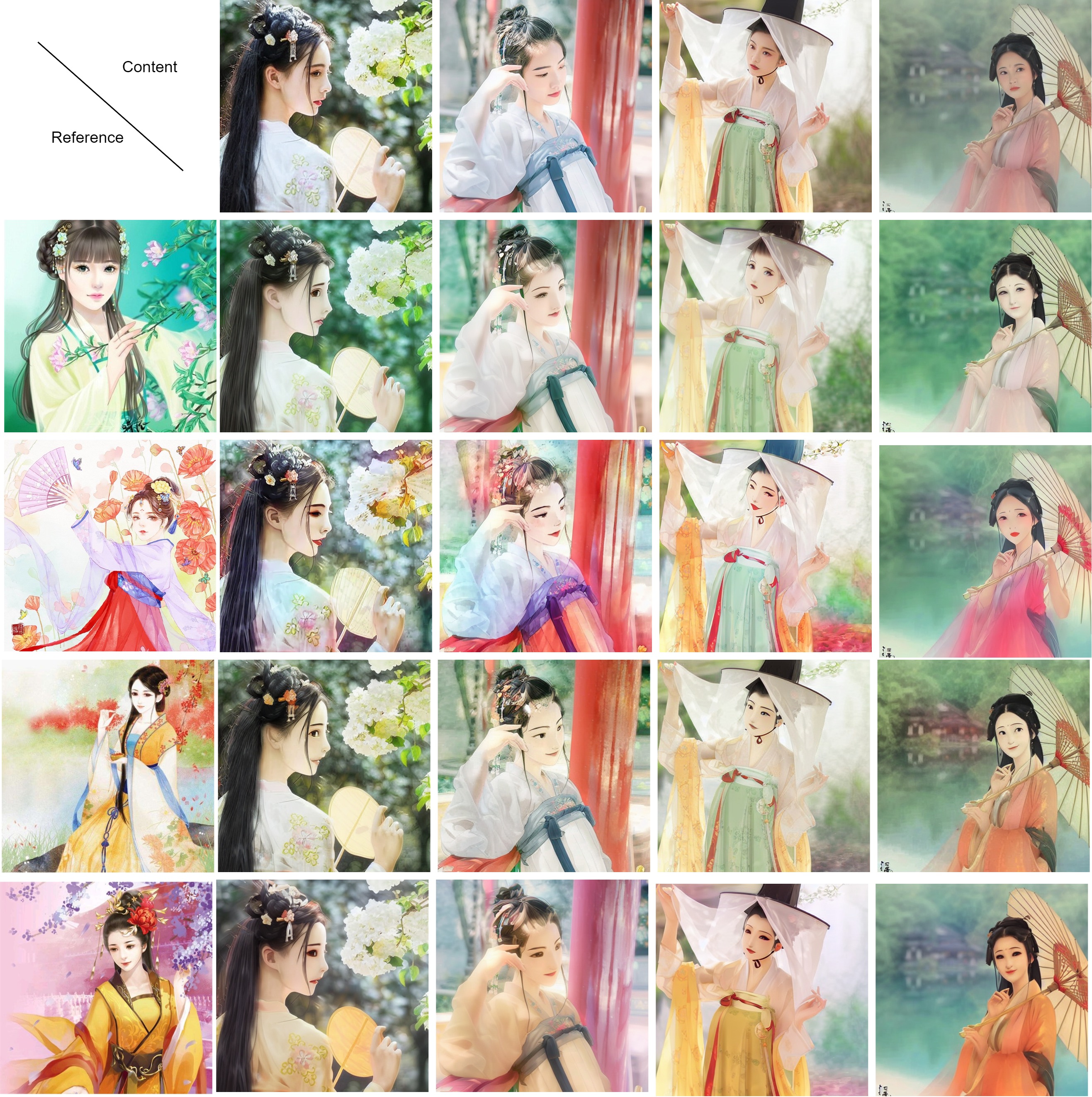}
\end{center}
\caption{ Model performance in style transfer tasks. The first column contains the reference images, and the first row contains the content images. The other images are the model outputs based on corresponding content and reference images.
}
\label{appendixfig:style_transfer}
\end{figure*}

To further verify the model performance in the style transfer task, we make more experiments with different reference styles, as shown in Fig.\;\ref{appendixfig:style_transfer}. In the figure, the first column contains the reference images, and the first row contains the content images. The model outputs show the excellent performance of the proposed method with content preserved and style transferred.

\begin{figure*}[t]
\begin{center}
\includegraphics[width=\linewidth]{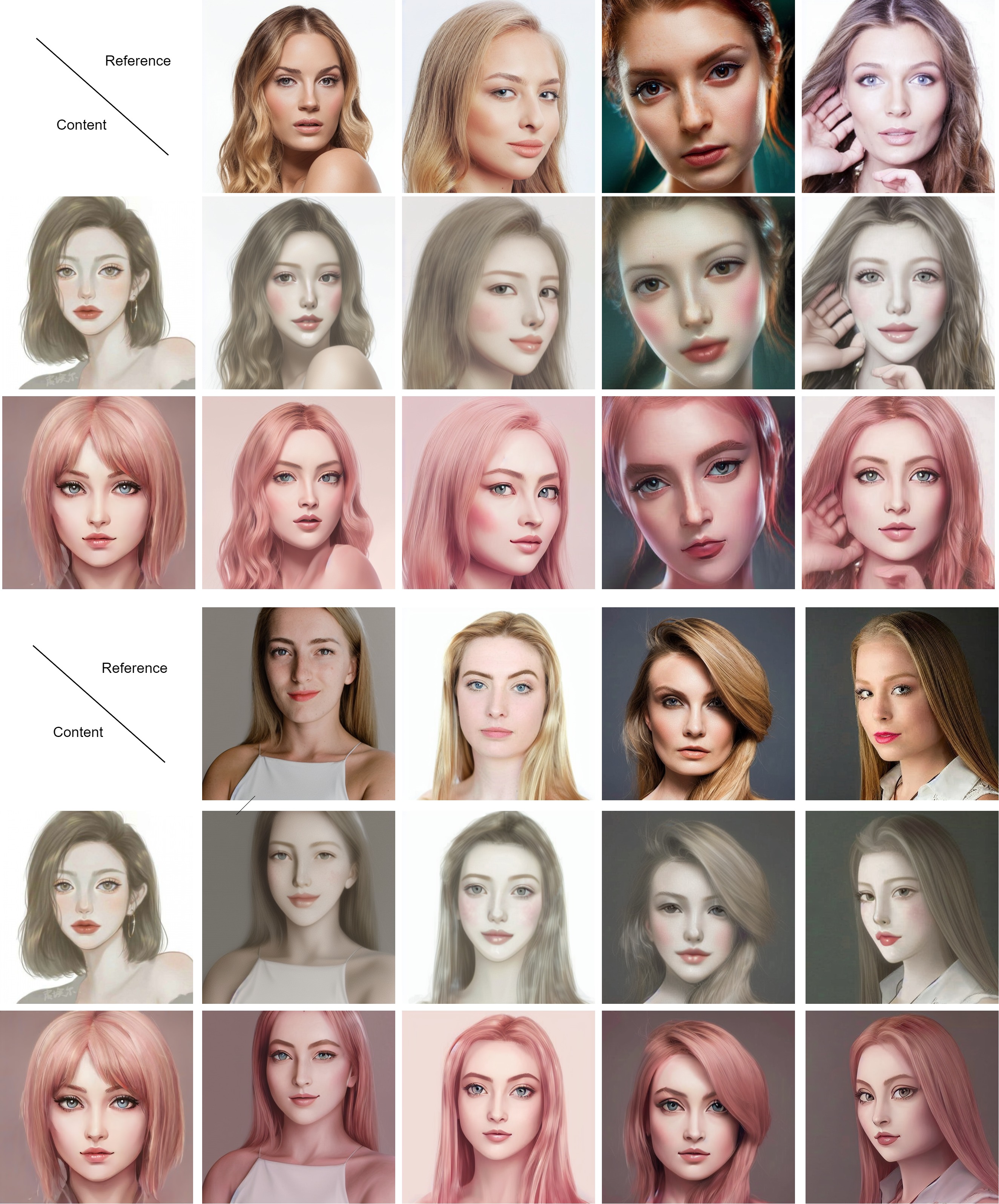}
\end{center}
\caption{ The model performance on portrait style transfer. Given the one-shot input, our model can substitute the face in the reference style image with the face in the content image with high quality.
}
\label{appendixfig:face_swap}
\end{figure*}

As a type of style transfer, portrait style transfer tries to substitute the input face with another stylized face. The proposed algorithm shows excellent performance in the task of portrait style as Fig.\;\ref{appendixfig:face_swap}. Given the one-shot input, our model can substitute the face in the reference style with the face in the content image with high quality.

\section{More Ablation Study}
\begin{figure*}[t]
\begin{center}
\includegraphics[width=\linewidth]{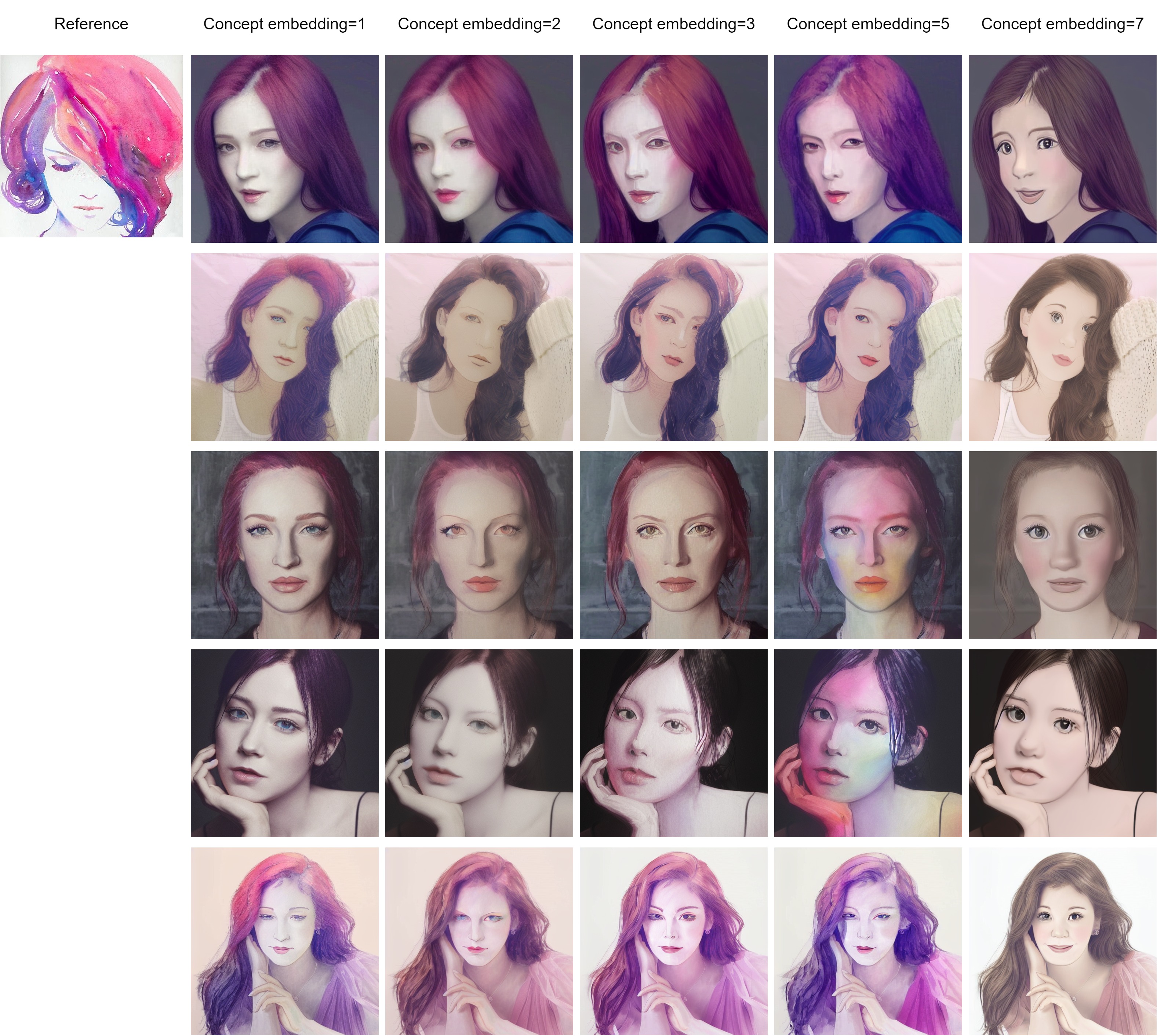}
\end{center}
\caption{ The model performance with different numbers of concept embedding. A small embedding number cannot well translate the concepts in the reference image, as in columns 2-3. A too-large embedding number still leads to poor performance with translation failures, as in columns 5-6. We empirically found that using 3 concept embeddings is the best choice, as in column 3.
}
\label{appendixfig:ablation_study}
\end{figure*}

We evaluate the influence of the number of multi-concept embeddings, as described in part of multi-concept inversion (Section 3.4) in the main paper. Given a reference image, we visualize the model performance with different concept embeddings as shown in Fig.\;\ref{appendixfig:ablation_study}. From the figure,  a small embedding number cannot well translate the concepts in the reference image, as in columns 2-3. A too-large embedding number still leads to poor performance with translation failures, as in columns 5-6. We empirically found that using 3 concept embeddings is the best choice, as in column 4.

\end{document}